\documentclass[graybox]{svmult}

\usepackage{type1cm}        
\usepackage{makeidx}         
\usepackage{graphicx}        
\usepackage{multicol}        
\usepackage[bottom]{footmisc}

\usepackage{newtxtext}       %
\usepackage{newtxmath}       

\usepackage{listings}
\usepackage{multirow}
\usepackage{soul}
\usepackage{wrapfig}
\usepackage{caption}

\makeindex             

\definecolor{gr}{rgb}{0.0, 0.5, 0.0}
\definecolor{b}{rgb}{0.0, 0.0, 1.0}
\definecolor{lb}{rgb}{0.0, 0.0, 0.5}
\definecolor{purple}{rgb}{0.55, 0.0, 0.55}
\definecolor{eclipseBlue}{RGB}{42,0.0,255}
\definecolor{eclipseGreen}{RGB}{63,127,95}
\definecolor{eclipsePurple}{RGB}{127,0,85}
\definecolor{lbcolor}{rgb}{0.9,0.9,0.9}  
\definecolor{lightblue}{rgb}{0.68, 0.85, 0.9}

\lstdefinelanguage{SPARQL}{
  showstringspaces=false,
  comment=[l]{//},
  morecomment=[s]{/*}{*/},
  commentstyle=\color{gr}\ttfamily,
  morecomment=[l]{\#},       
  morecomment=[n][\color{lb}]{<http}{>}, 
  morestring=[b][\color{green}]{\"},  
  classoffset=1,
  morekeywords={?O, sh, sosa, owl, xsd, purl, ssr, ssn}, keywordstyle=\color{b},
  classoffset=2,
  morekeywords={
    SELECT,CONSTRUCT,DESCRIBE,ASK,WHERE,FROM,NAMED,PREFIX,BASE,OPTIONAL,
    FILTER,GRAPH,LIMIT,OFFSET,SERVICE,UNION,EXISTS,NOT,BINDINGS,MINUS,
    STREAM, NAF, WINDOW, RANGE, ON
  },
  keywordstyle=\color{gr}\textbf,
  classoffset=3,
  morekeywords={
   det, det1, det2, det3, det4, det5,
   car, hasConfScore, 
   resultTime, isSampleOf, hasResult, usedProcedure, Detection, Box3D, isIn, nearby,
   b1,b2,b3,b4,b5,b6,b7,
   trk,trk23,trk5,
   rule_w_1,rule_w_2, rule_w_3, rule, leaves, inFOV, rule_w_4, vMatch, ends,
   NodeShape, FoV, CQELSRule, prefixes, construct, rule_w_5,
   enters, trklet
  },
  keywordstyle=\color{purple},
  classoffset=4,
  otherkeywords={&&},
  morekeywords={
    window, sec, iou, @
  },
  keywordstyle=\color{red},
  classoffset=5,
  morekeywords={a}, keywordstyle=\color{lb}\textbf,
}

\newcommand\nop[1]{}

\begin{document}

\title*{SemRob: Towards Semantic Stream Reasoning for Robotic Operating Systems}

\author{Manh Nguyen-Duc, Anh Le-Tuan, Manfred Hauswirth, David Bowden and Danh Le-Phuoc}
\authorrunning{Manh Nguyen-Duc et la. } 
\institute{Manh Nguyen-Duc \and Anh Le-Tuan \at Open Distributed Systems, Technical University of Berlin, Germany
\and David Bowden \at Dell Technologies, Ireland
\and Manfred Hauswirth \and Danh Le-Phuoc \at Open Distributed Systems, Technical University of Berlin\newline Fraunhofer Institute for Open Communication Systems, Berlin, Germany}

%
%
\maketitle


\abstract{Stream processing and reasoning is getting considerable attention in various application domains such as IoT, Industry IoT and Smart Cities.  In parallel, reasoning and knowledge-based features have attracted research into many areas of robotics, such as robotic mapping, perception and interaction. To this end, the Semantic Stream Reasoning (SSR) framework can unify the representations of symbolic/semantic streams with deep neural networks, to integrate high-dimensional data streams, such as video streams and LiDAR point clouds, with traditional graph or relational stream data. As such, this positioning and system paper will outline our approach to build a platform to facilitate semantic stream reasoning capabilities on a robotic operating system called SemRob.}

\section{Motivation}
\label{sec:1}

Semantic stream processing and reasoning are getting more and more attention in various application domains such as IoT, Industry IoT, and Smart Cities~\cite{Le-Phuoc:2010,Daniel:2016,Daniel:2017,Linda:2019,Manh:2019}. Among them,  many recent works, e.g.~\cite{Suchan:2019,Le-Phuoc:2021}, were motivated by several use cases in autonomous driving and robotics. In parallel, reasoning and knowledge-based features attracted many research in robotics such as KnowRob~\cite{Michael:2018} and semantics for robotic mapping, perception, and interaction~\cite{Sourav:2020}. Moreover, various works, e.g~\cite{Lyu:2019,Illanes:2020}, on symbolic planning for robotics  have a lot of connections to semantic streams and reasoning.

Along this line, the semantic stream reasoning (SSR) framework~\cite{Le-Phuoc:2021} unifies the representation of symbolic/semantic streams~\cite{Le-Phuoc2018-b} with deep neural networks (DNNs) to integrate high-dimensional data streams, such as video streams and LiDAR point clouds, with traditional RDF graph streams. To this end, SSR framework paves the roadway to use its semantic stream representation and its probabilistic reasoning operations to abstract data processing pipelines and interactions of the Robotic Operating Systems (ROS).

Hence, this motivated us to propose a declarative programming model for ROS via SSR. This programming model will be supported by a reasoning kernel, called SemRob agent, which marries ROS abstractions with SSR presentations and processing primitives. This paper will present how we will release SemRob with the following sections. The next section will present how we align SSR with design abstractions of ROS. Section~\ref{sec:arch} will propose an architecture to realize SemRob. Next, Section~\ref{sec:app} will show how to realize three typical application scenarios in ROS with SemRob. Finally, we will outline our implementation roadmap and some demonstration ideas to evaluate the SemRob platform.

\section{Semantic Stream Reasoning Towards ROS-based Abstractions}
\label{sec:ssr}
SSR framework enables data fusion from multi-modal stream data, such as camera and LiDARs, via declarative rules. Such rules can be written in SPARQL-like or ASP syntaxes~\cite{lifschitz:2019}. To this end, the stream data flowing among data fusion operations can be represented as standardized data formats, e.g., RDF-star. This enabled us to build a federated processing architecture, e.g.  for distributed video streams in ~\cite{Manh:2021}. In the architecture of this kind, an autonomous stream fusion agent  can be modeled as  a unified abstraction of a processing node or sensor source, e.g a camera or LiDAR. This abstraction is aligned with the basic abstraction in ROS design, namely \textbf{ROS Graph}~\cite{quigley2009ros}. ROS Graph is a network of multiple nodes which communicate with each other to exchange data. Each node in ROS should be responsible for a single module purpose, e.g. one node for controlling wheel motors, one node for controlling a laser range-finder, etc. Each node can send and receive data to other nodes via topics, services or actions.

\begin{figure}[ht!]
\vspace{-3mm}
    \centering
    \includegraphics[width=0.75\textwidth]{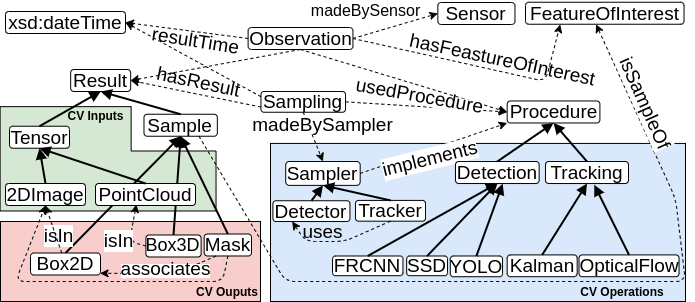}
    \caption{Abstraction of Sensory Streams with SSN}
    \label{fig:ssn}
\vspace*{-1.5\baselineskip}   
\end{figure}

Our design, in the next section, will focus on ROS 2~\cite{ros2:2018}, which is a middle-ware based on an anonymous publish/subscribe mechanism that allows for message passing between different ROS processes.  To align the ROS node abstraction with SSR representation, we can follow the neural-symbolic stream modeling practice in~\cite{Le-Phuoc:2021} to extend the standardized Semantic Sensor Network Ontology (SSN)~\cite{Armin:2019} to capture sensor readings, communication messages and intermediate processing states across ROS nodes. Such an extension can include various vocabularies to specify the semantics of camera sensors, video frames, and tensors  as shown in Figure~\ref{fig:ssn}.

 Note that using RDF-based symbols from such vocabularies can facilitate semantic interoperability across distributed ROS nodes, called \emph{Semantic Nodes}. With the declarative continuous query language~\cite{lephuoc:2011,Daniele:2016} ,  the query federation feature  in~\cite{Manh:2019} and~\cite{Manh:2021} is aligned with the publish/subscribe mechanism along with distributed data distribution abstraction, such as Data Distribution Service (DDS) of ROS. For example, the following CQELS-QL query in Listing~\ref{lst:fuse2sensors} will subscribe to a continuous query to fuse two videos, and publish a new stream as a ROS node, generating bounding boxes of "traffic obstacles" utilizing object detection models.

\begin{lstlisting}[
caption={Fuse two camera with CQELS-QL}, 
label={lst:fuse2sensors},
language=SPARQL, 
captionpos=b,
basicstyle=\footnotesize\ttfamily,  
backgroundcolor=\color{lbcolor},
numbers=left,
escapeinside=||,]
REGISTER <:leftright2DBoxes> AS
CONSTRUCT {
<<?lbox  :frontLeftOf ?veh> ssn:resultTime ?time>>.
<<?rbox  :frontRightOf ?veh> ssn:resultTime ?time>>.
}
WHERE {
  STREAM {:frontLeftCamNode} [RANGE 5s ON ssn:resultTime] {
    ?lBox  a                  ssr:TrafficObstacle.
    ?obs ssn:hasResult ?box.
    ?obs ssn:resultTime ?time.
  }
  STREAM {:frontRightCamNode} [RANGE 5s ON ssn:resultTime] {
    ?rBox  a                  ssr:TrafficObstacle.
    ?obs ssn:hasResult ?rBox.
    ?obs ssn:resultTime ?time.
  }
  :frontLeftCamNode :generatedBy ?leftCam.
  ?leftCam :mountedOnFrontLeft ?veh.
  :frontRightCamNode :generatedBy ?rightCam.
  ?rightCam :mountedOnFrontRight ?veh.
}
\end{lstlisting}

Thanks to the reasoning capability of SSR, traffic objects are semantically inferred from the knowledge graph containing an ontology of cars, trucks, and traffic lights.  In this example, the URL \textit{:frontLeftCamNode} is used to identify the identity of the node that handles data generated from the left camera (line 7), and the same for the URL \textit{:frontRightCamNode} is used to identify the identity of the node that handles data generated from the right camera (line 12). In this query, we want to get the bounding box of traffic obstacles in the last 5 seconds from two different cameras on the vehicle. 




The output variables, ?veh and ?time, can be used to correlate with GPS coordinates to localize the obstacles by projecting the bounding boxes, i.e., ?lbox and ?rbox, to the vehicle's GPS coordinate system with standardized semantic spatial representations as suggested in~\cite{Linda:2019}. This can later be used for the Semantic SLAM use cases in section~\ref{sec:app}.

\nop{
\begin{lstlisting}[
caption={SHACL rules with CQELS-QL}, 
label={rule1},
language=SPARQL, 
captionpos=b,
basicstyle=\footnotesize\ttfamily,  
backgroundcolor=\color{lbcolor},
numbers=left,
escapeinside=||,]
SELECT ?box3d ?timeStamp
WHERE {
  STREAM ?carURI [RANGE 5m ON ssn:resultTime] {
    ?observation   ssn:usedProcedure  ?procedure.
    ?procedure     a                  ssr:Detection.
    ?observation   ssn:resultTime     ?time.
    ?observation   ssn:hasResult      ?result.
    ?result        a                  ssr:Box3D.
    ?result        ssr:isIn           ?pointCloud.
  }
  ?carURI    :nearby     ?loc.
}
\end{lstlisting}
}

\begin{figure}[ht!]
\vspace{-4mm}
    \centering
    \includegraphics[scale=0.085]{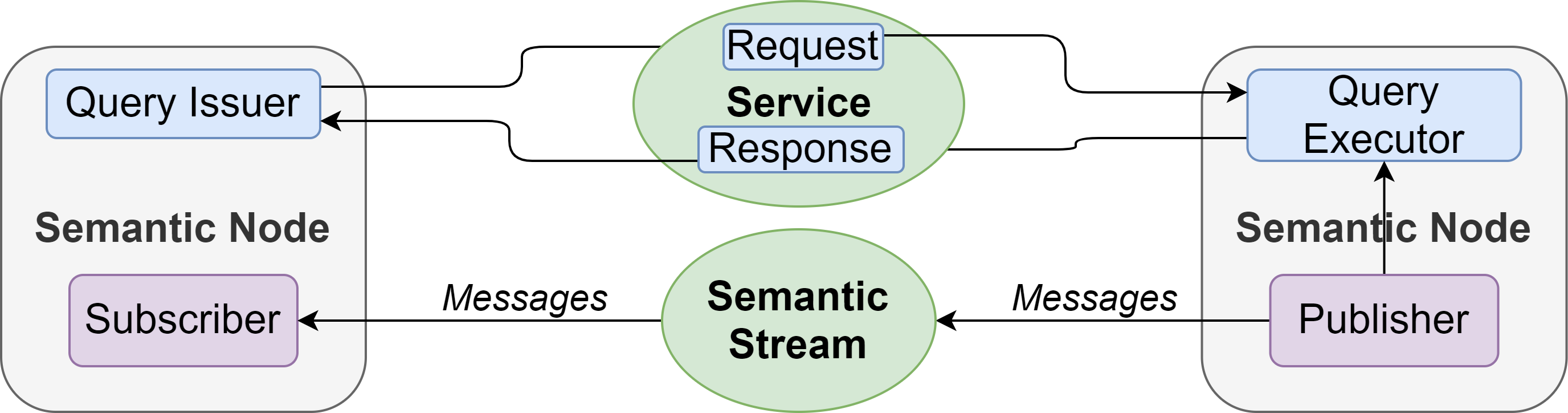}
    \caption{Communicate between two Semantic Nodes}
    \label{fig:ros2}
    \vspace*{-2\baselineskip}   
\end{figure}

The Semantic Nodes of ROS2 communicate using a \textit{Service} to send the request and receive the response (Figure~\ref{fig:ros2}). To exchange data, a node uses a publisher to create a \textit{Semantic Stream} (a.k.a a Topic in ROS2) and send data as \textit{Semantic messages} encoded in a ROS2 standardized message format, such as JSON-LD or an RDF-star serialization format, to that  stream. Another node in the network, which subscribed to that semantic stream, will receive the data once the message has been sent from the publisher. A node could have both \textit{Publisher} and \textit{Subscriber} to send and receive the message. Note that, this communication can happen within one SemRob agent or two SemRob agents, connected via wireless networks. 

\section{Architecture Design of SemRob}
\label{sec:arch}

\begin{figure}
    \vspace{-1.5mm}
  \begin{center}
    \includegraphics[scale=0.078]{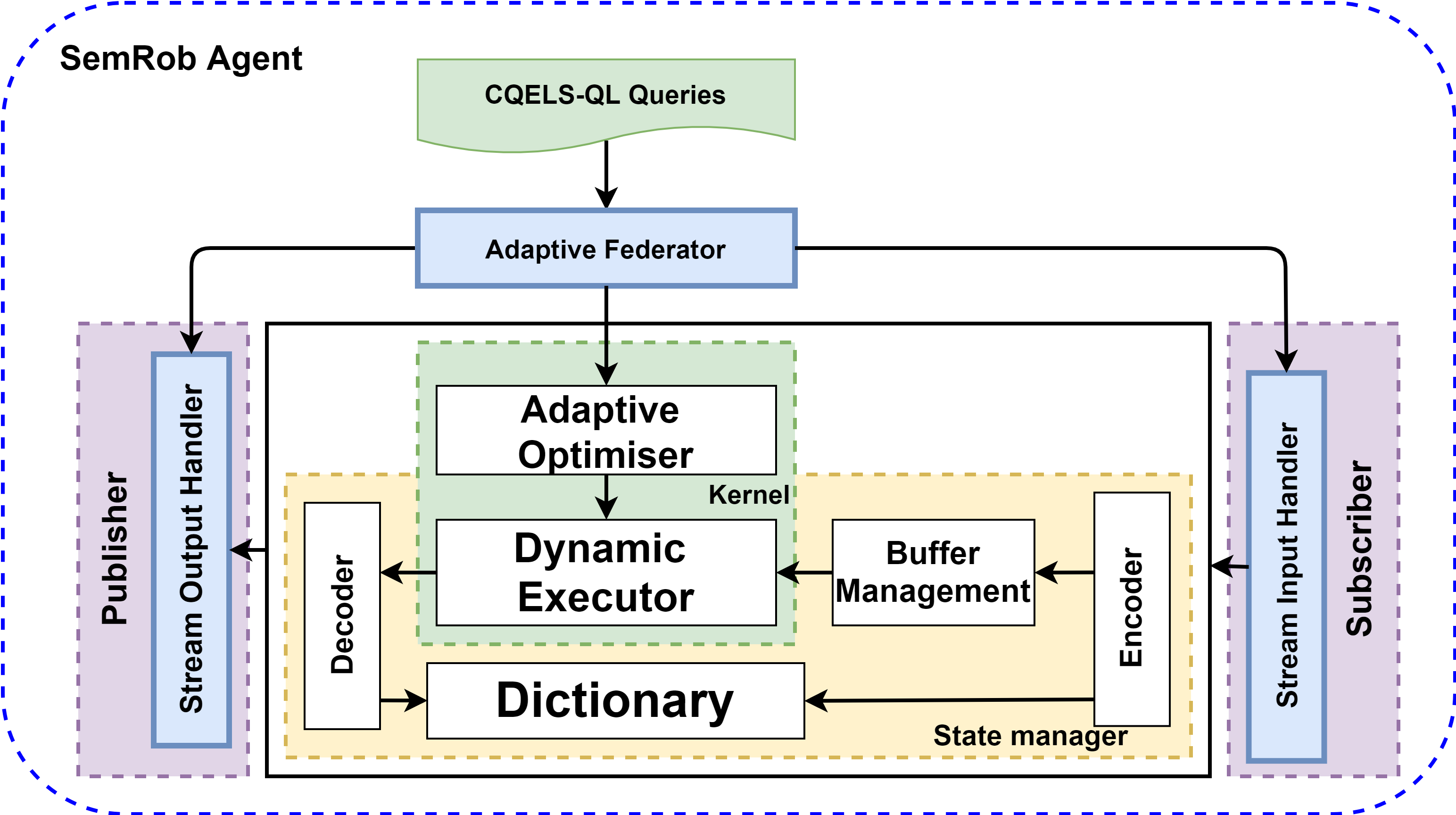}
  \end{center}
  \caption{Architecture of SemRob Agent}
  \label{fig:architecture}
  \vspace*{-2\baselineskip}   
\end{figure}

To build the SemRob platform on top of ROS2, our architecture design is illustrated in Figure~\ref{fig:architecture}. This design extends the Autonomous Fusion Agent (ASF) from Fed4Edge~\cite{Manh:2021} with the communication and APIs supported by ROS2. In particular, the topic-based publish/subscribe mechanism of ROS is integrated with the input/output handlers for the ASF agent, which serves an important role in communicating between nodes. The core is the processing kernel and the \textit{State manager}, which reuses the processing components of Fed4Edge and SSR. The \textit{Adaptive Federator} will help to transparently coordinate between SemRob agents as a network of semantic nodes, which abstracts away the network communications. 

Thanks to the semantic abstraction of SSR, integrating more sensors into a data fusion pipeline can be done at run-time. For example, Listing~\ref{fuse3sensors} can use the output from Listing~\ref{lst:fuse2sensors} above to add a LiDAR sensor into its fusion pipeline to create a new one. Lines 10 to 14 will add a stream of LiDAR data presenting to the ROS node, extracting bounding boxes from the LiDAR sensor mounted on the same car as the two camera sensors.

ROS2 supports a distributed discovery method that can be used to handle the publish and subscribe topics in a decentralized fashion, which abstracts away the communication and network boundaries. When a node goes online or offline, it will advertise its information to all other existing nodes in the same ROS domain (ROS domain is pre-configured in environment variables) by using \textit{Data Distribution Service} (DDS) publish-subscribe transport. Once nodes in the system receive the advertisement, they will send a response with their information so they could communicate with each other.
\begin{lstlisting}[
caption={Fuse three sensors with CQELS-QL}, 
label={fuse3sensors},
language=SPARQL, 
captionpos=b,
basicstyle=\footnotesize\ttfamily,  
backgroundcolor=\color{lbcolor},
numbers=left,
escapeinside=||,]
REGISTER <:boxesNode> AS
CONSTRUCT {
    <<?box2d  :frontOf ?veh> ssn:resultTime ?time>>.
    <<??box3d :fromLidarViewOf ?veh> ssn:resultTime ?time>>.
}
WHERE {
  STREAM {:leftright2DBoxes} [RANGE 5s ON ssn:resultTime] {
    <<?box2d  :frontOf  ?veh> ssn:resultTime ?time>
  }
  STREAM {:lidarNode} [RANGE 5s ON ssn:resultTime] {
    ?box3d  a                  ssr:TrafficObstable.
    ?obs ssn:hasResult ?box3d.
    ?obs ssn:resultTime ?time.
  }
  :lidarNode :generatedBy ?lidar.
  ?lidar :mountedOnOnTop ?veh.
}
\end{lstlisting}

\begin{lstlisting}[
caption={Dynamic Discovery with CQELS-QL}, 
label={lst:fusewithdiscovery},
language=SPARQL, 
captionpos=b,
basicstyle=\footnotesize\ttfamily,  
backgroundcolor=\color{lbcolor},
numbers=left,
escapeinside=||,]
SELECT ?box ?sensor ?time
WHERE {
  STREAM ?stream [RANGE 5s ON ssn:resultTime] {
    ?box  a                  ssr:TrafficObstable.
    ?obs sosa:hasResult ?box.
    ?obs ssn:resultTime ?time.
  }
  ?stream :generatedBy ?sensor.
  ?sensor :mountedOn :myCar.
}
\end{lstlisting}

This feature allows us to enable the dynamic discovery of all available sensor streams without having to specify the URIs of relevant streams. Listing~\ref{lst:fusewithdiscovery} shows an example that fuses all streams based on the matched metadata, e.g, lines 8-9. Note that \emph{:mountedOn} is a super-property of the property \emph{:mountedOnFrontLeft},\emph{:mountedOnFrontRight} and \emph{:mountedOnTop} in the Listing~\ref{fuse3sensors}. Thanks to the reasoning capability of SSR, the reasoning engine will infer such matched sub-properties of \emph{:mountedOn}. In this way, the SemRob agent or a semantic node can join the network without interrupting the on-going processing pipelines or interactions.

\section{Application scenarios}
\label{sec:app}
We target three typical application scenarios that are interwoven in terms of feature dependencies and data flow hierarchies. These scenarios will drive our decisions in the architecture and system design of Section~\ref{sec:5} below. They also provide evaluation scenarios to show the advantages of SemRob against traditional approaches.

\subsection{Declarative Multi-sensor fusions}

Multi-source and heterogeneous information fusion~\cite{Wang:2020}  makes integrated utilization of the information obtained by different sensors, which avoids the perceptual limitations and uncertainties of a single sensor, forms a more comprehensive perception and recognition of the environment or target, and improves the external perception ability of the system. 
Various sensors have their advantages and disadvantages, so do their derived perceptual tasks. In particular, the camera can acquire the optical image and accurately record the contour, texture, color distribution, and other information of the object from a certain angle. Therefore, some studies use cameras for perceptual tasks such as 
lane detection, pedestrian and vehicle identification, and local path planning, e.g.~\cite{Pendleton:2017}.
It can maximize the reconstruction of traffic conditions in the real-world environment to combine the dynamic characteristics of the MMW-Radar targets, the ranging advantage of the LiDAR, and the details of the target in the optical image.
Appropriate utilization of integrated information facilitates vehicles to perform diverse tasks such as intention analysis, motion planning, and autonomous driving. Different from the previous imperative program approach in multisensor fusions which need prior knowledge of input sensors, SemRob's declarative approach as shown in the Listing~\ref{lst:fusewithdiscovery} can integrate new sensor sources in real-time. Note that  similar work like~\cite{Daniel:2016,Daniel:2017} also proposed to use semantic stream data in ROS,  however, it is not clear how multimodal sensor streams along with DNNs will be represented and processed in a uniform way like our approach.


\subsection{Semantic SLAM}
Simultaneous Localization And Mapping (\textbf{SLAM}) in robotics is a process that the robot or the autonomous vehicle used to create a map of the surrounding environment. While building the map, the robots also locate its location and orientation in the map\cite{slam}, then using its generated map to explore the areas and continue to generate the new map. \textbf{Semantic SLAM} is a process that creates a map that has both \textit{metric} information (such as location and orientation) and \textit{semantic} information about objects or landmarks in the map by applying machine learning methods~\cite{Dong:2020,Shuhuan:2021} 

Based on the ability to fuse sensor data from multiple sources in the above section, SemRob can help build Semantic SLAM more easily and effectively. Semantic SLAM builds a map by fusing data from multiple sensors, such as cameras for taking surrounding environment vision and 2D object detection, LiDAR for 3D object detection and measuring distance, and IMU for localization and orientation. While building the map, it  also detects objects such as cars and pedestrians, which are part of the surrounding environment, and then classifies and gives them semantic labels.
For example, we use the SSN vocabulary as the Figure~\ref{fig:ssn} for presenting data, and the query in Listing~\ref{lst:fusewithdiscovery} for gathering data from sensors and devices. By using the query in Listing~\ref{lst:fusewithdiscovery} to subscribe to camera streams, LiDAR stream, and IMU stream, the combined streams can be used to process and build the map.

\begin{figure}
    \vspace{-5mm}
    \includegraphics[scale=0.6]{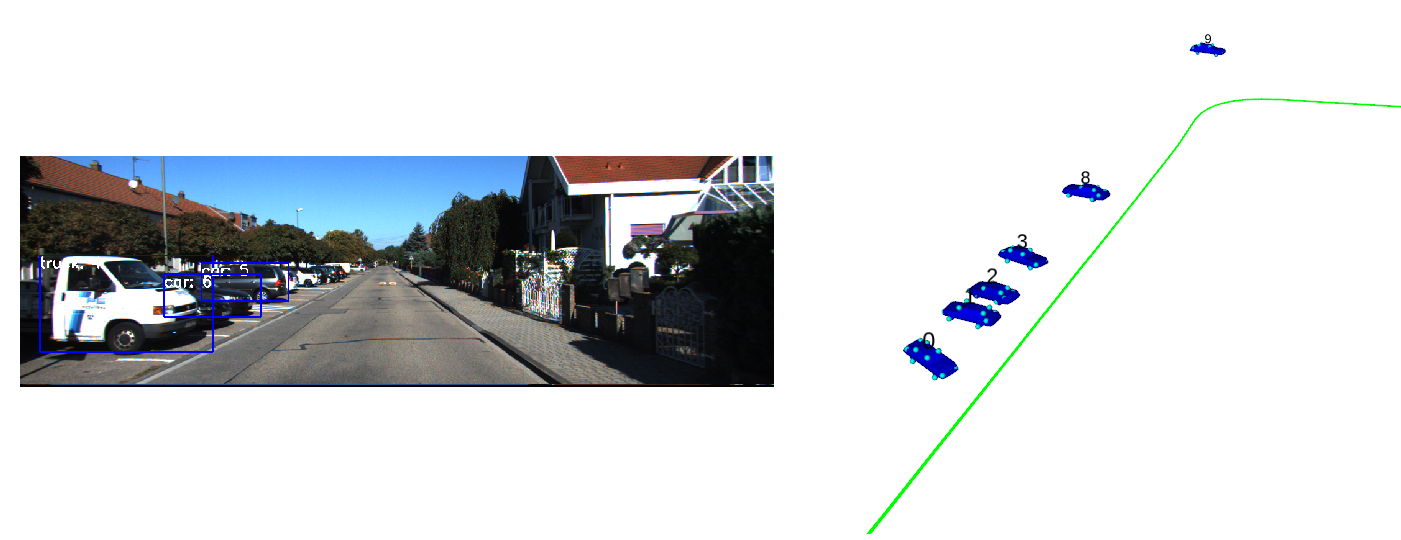}
    \caption{Sample image from the series of images in KITTI dataset (left) and Semantic SLAM map (right) shows trajectory and objects (cars) in the scene. \cite{slam}}
    \label{fig:sslam}
    \vspace*{-1\baselineskip}   
\end{figure}

Semantic SLAM enables human-robot communication through words that describe objects or landmarks on the map. It facilitates reasoning about the scene for autonomous vehicles that have to make decisions based on “what” they see. Lastly, they provide even better metric maps by abstracting objects with their geometric shapes. Hence, the reasoning capability powered by SSR will provide SemRob advantage in use cases of this kind.

\subsection{Cooperative Perception}

Cooperative perception facilitates the possibility to enhance the overall perception by aggregating information received from multiple nearby vehicles. In particular, it is  considered as a 'killer application' for cooperative driving in the context of 5G  for traffic safety and efficiency~\cite{Boccardi:2014,Gui:2020,Viswanathan:2020},  because  the  resultant merged map extends the perception range beyond line-of-sight and sensing angles. Accordingly, they enable 
early obstacle  detection and long-term  perspective  path  planning~\cite{Kim:2015-hs}.

To date, most of the recent works have been focusing only on one of the following aspects: how to encode the communication message, how to communicate, and how to integrate data from a known set of remote sensing sources. In particular, some work has utilized messages encoding three types of data: raw sensor data, output detection, or metadata messages that contain vehicle information, such as location, heading, and speed. Rauch et.al,~\cite{Rauch:2017} associate the received V2V messages with the outputs of local sensors. The works like ~\cite{Chen:2019,Arnold:2020} propose to aggregate LiDAR point clouds from other vehicles, followed by a deep network for detection. Rawashdeh et.al,~\cite{Rawashdeh:2018} process sensor measurements via a deep network and then generated perception outputs for cross-vehicle data sharing. Wang et.al,~\cite{Tsun-Hsuan:2020} leverage V2V communication to improve the perception and motion forecasting performance of self-driving vehicles using LiDAR and cameras. 
While such approaches can be seen as offering different convincing evidence, showing the advantage of having robust and scalable cooperative perception solutions, most of them only assume the vehicles have a fixed set of sensors and the same processing pipelines, e.g. same DNN models. Hence, we aim to overcome such problems of dynamic integration and abstraction of sensor sources and processing states in  mobile networks of vehicles using SemRob. We are not the first try to address this interesting problem using the knowledge representation approach. In fact, some work such as KnowRob~\cite{Moritz:2013,Michael:2018} and  DyKnow~\cite{Daniel:2016} touch the cooperative perception problem in the similar direction as ours. But we believe that the problem is still very difficult and widely open according to various recent survey and empirical studies across the many fields such as ~\cite{Rauch:2017,Chen:2019,Gui:2020,Viswanathan:2020,Arnold:2020,Tsun-Hsuan:2020,Mengtian:2020}.

\section{Implementation Roadmap and Demonstration ideas}
\label{sec:5}

To evaluate our targeted SemRob platform, we are building a testbed for setting up  networks of robots that can cooperatively carry out the tasks mentioned in Section~\ref{sec:app}. To enable iterative evaluations  and hierarchical learning schemes in the development phases, networks will be equipped with sensors, wireless connections and embedded computing nodes to emulate a fleet of cars (e.g. for platooning \cite{Caveney:2012} or convoy formations~\cite{Sawade:2015}) or drone fleets. The testbed will be designed following a modular and open-source design such as~\cite{Matthew:2019,Keylly:2020} so that we can test different combinations of sensing, networking, and computation capabilities of these vehicles.

In particular, one of the demonstrations under development is to build an autonomous robot-car network that can share the sensing capability in real-time via CQELS queries. Next is  to build a cooperative Semantic SLAM map via SemRob. This system can enable cars and drones to cooperatively build and share real-time semantic maps towards cooperative perception discussed in~\cite{Boccardi:2014,Gui:2020,Viswanathan:2020}. Another demonstration planned is to build a collaborative planner that can coordinate actions of robotics by incorporating stream from multi-context framework~\cite{Minh:2017} with SSR.

\begin{itemize}
    \item \textbf{Autonomous Robot-Car Network:} We are going to build a fleet of cars which have different kinds of sensors. The idea behind this is that they could share and reuse the data captured and processed from their peers instead of capturing and processing it themselves. Based on the federation mechanism from \cite{Manh:2019} and \cite{Manh:2021}, all cars in the network will communicate and exchange data with each other through CQELS-QL queries.  Moreover, a robot-car could take the data from the others around them to extend its view or improve its prediction. 
    \item \textbf{Cooperative Semantic SLAM:} The idea is that multiple robots would take different routes over the same map, then they will construct that map together in real-time. Each robot will be responsible for different parts of the map. This scenario will need the features built from the previous demonstration scenario. 
   \item \textbf{Collaborative Planner:} This scenario will focus on the robot planning problem with collaborative functionality. In this scenario, we add drones into the network that will follow the cars to give the bird's eye view data. With the advantage of the wide view from above, the system could have a general picture about the current traffic to make plans for several scenarios. However, each vehicle has its operating constraints such as sensing range, battery life, and processing power, therefore we need to build a collaborative planer to coordinate the actions of the vehicles to accomplish a certain task (e.g the previous two tasks) according to specified optimization goals, such as saving battery life or minimizing prediction errors.
\end{itemize}

\begin{figure}
\vspace{-4mm}
\centering
\begin{minipage}{.5\textwidth}
\centering
    \includegraphics[scale=0.08]{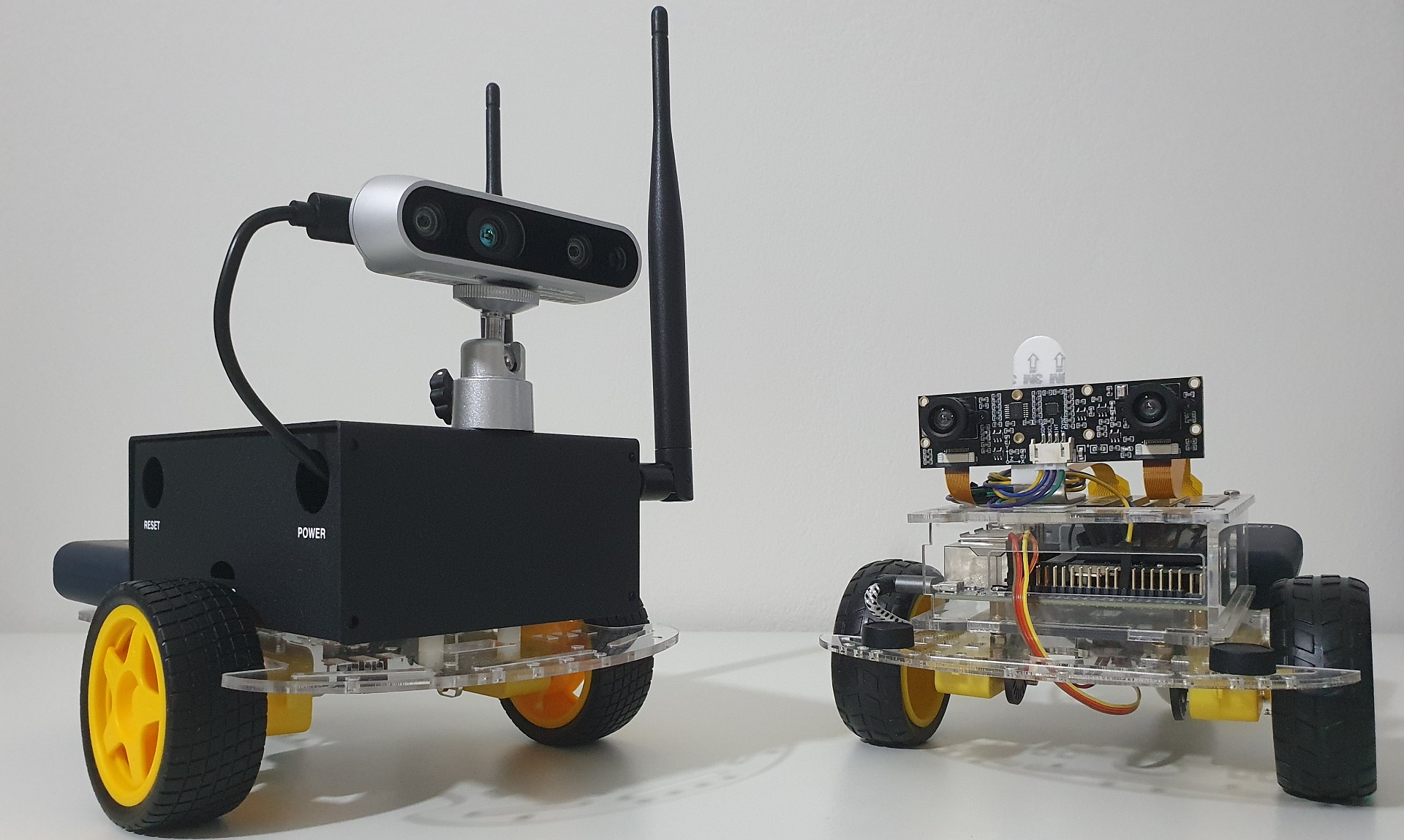}
    \caption{Hardware setup for small-scale self-driving cars/robots}
    \label{fig:demo}
\end{minipage}%
\begin{minipage}{.5\textwidth}
\centering
    \includegraphics[scale=0.08]{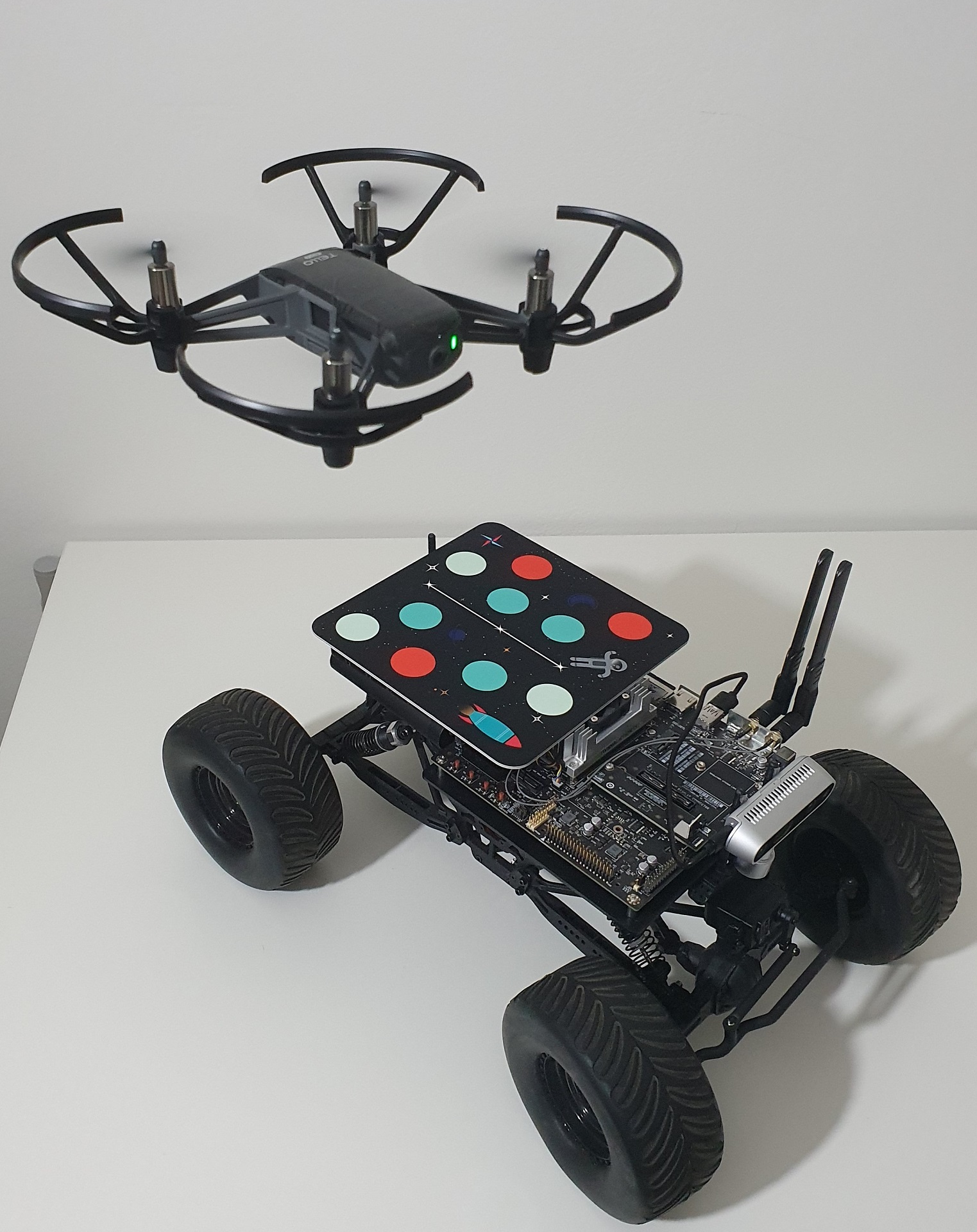}
    \caption{Hardware setup for large-scale self-driving cars/robots and drone network}
    \label{fig:demo2}
    \end{minipage}
\vspace*{-1\baselineskip}   
\end{figure}

For the robot-car models, we are building a fleet of three types of model robot-car that have different hardware configurations, as illustrated in Figure~\ref{fig:demo}:
\begin{itemize}
    \item \textbf{Model 1:} This is the basic model assembled in a small robot-car chassis with the lowest computational processor. This model is equipped with a Jetson Nano computer board, which has a Quad-core ARM Cortex-A57 MPCore processor CPU, 4GB RAM, a GPU NVIDIA Maxwell with 128 NVIDIA CUDA cores. The Jetson Nano could run DNN detection models for simple tasks on data streams from a mono or stereo camera. This model is used to emulate small robots on factory floors or harvesting farms.
    \item \textbf{Model 2:} This model is assembled with a medium-size robot car chassis. It is equipped with a Jetson TX2/NX computer board, which has medium computational power. For example, the Jetson TX2 has Dual-Core NVIDIA Denver 2 64-Bit CPU, Quad-Core ARM Cortex-A57 MPCore, 8GB RAM, and 256-core NVIDIA Pascal™ GPU with 256 NVIDIA CUDA cores. This board has twice as much processing power and memory compared with \textit{model 1}, so it can play the role of a gateway to the internet/WAN, and it can be used to offload processing from a set of weaker Model 1 robot-cars. Since it has more computational power, it could be equipped with a stereo camera with IMU or an Intel Realsense 3D camera for 3D processing, such as 3D object detection, or cooperate to create a Semantic SLAM map. Due to the increased computational processing power, it will consume much more energy, therefore a 10,000 mAh battery is used as the power source.
    \item \textbf{Model 3:} This model is built on a large robot-car chassis so it can carry more devices and sensors. This model is equipped with a Jetson Xavier AGX computer board, which has 8-core ARM v8.2 64-bit CPU, 32GB RAM, and 512-core Volta GPU with Tensor Cores. This Jetson Xavier AGX has a similar capability to the computer installed in real self-driving vehicles. With this high processing power computer board, a robot-car of this type could perform and handle many heavy computing tasks. This model can be used to coordinate the actions of smaller robot-cars. Moreover, each could be mounted with a stereo camera or an Intel Realsense 3D camera for vision tasks, and it can even handle LiDAR sensors and processing point clouds with DNNs on board. This robot-car could act as a host for offload processing or cooperating with drones. The vehicle of this model will consume much more energy than the previous ones, so we use a high-capacity battery, e.g 25,600 mAh.
\end{itemize}

\textbf{Sensors with models:} Based on the computational power of each model, they could be equipped with several sensors/devices to perform a particular task in the robot network. Models 1 and 2 can be mounted with a mono or stereo camera, IMU sensor, to do DNN detection or cooperate to build a Semantic SLAM map. Model 2 and 3 could be equipped with a stereo camera or a 3D camera such as Inter Realsense 3D camera for increasing the accuracy of Semantic SLAM map or simulated as a pseudo-LiDAR for 3D object detection. Model 3 with its high computation power could handle heavier sensors, such as LiDARs, radars, and multiple cameras to perform the more complicated tasks, such as building a high-quality Semantic SLAM map or 3D object detection.

For the drones, we will also build a fleet with two different models. A remote-control model with a basic setup from the manufacturer and a custom built model with on-board control and different types of sensors, as illustrated in Figure~\ref{fig:demo2}.
\begin{itemize}
    \item \textbf{Remote-control model:} This model uses  Tello Edu drones which are equipped with an HD camera and autopilot functionality. Since this model does not have the on-board computational power,  it could be used for streaming video and could be  remotely controlled by a robot-car on the ground. The video stream can be sent from this drone; thus, it will be used in the tasks of \textit{Collaborative Planner}. This drone could activate the follow-mode with the autopilot control to follow a robot on the ground by mounting a \textit{Mission Pad} on top of the robot-car.
    \item \textbf{Onboard-control model:} In this model, we custom-build large drones with  F550 frames (each frame with the diameter of 550cm) using a 3D printer, each frame is printed with customized slots for mounting sensors and devices. This large drone could carry more sensors and devices for complicated tasks. It could carry a Jetson Nano computer board as described above as well as  a LiDAR for object detection tasks and collision avoidance, and a camera for vision tasks. The drone could work standalone or cooperatively with robots on the ground for many different tasks. The drone is powered by a 4,000 mAh battery for flying and compute operations.
\end{itemize}

\section{Conclusions}

This paper presents our position idea together with the research roadmap for the SemRob project. We presented our initial ideas with the first design of the research platform towards three interwoven application scenarios, namely, declarative multi-sensor fusion, semantic SLAM and cooperative perception. The aim is to build practical applications along with progressing the research; and we presented how we design a sophisticated testbed to evaluate the SemRob on realistic scenarios.

We believe that the proposed testbed described above, will allow us to fully investigate the potential of SemRob. The strategy of layering streaming sensor fusion on top of ROS, which can then be used to extend SLAM with the concept of Semantic SLAM, is both novel and will elaborate on existing concepts. Additionally, the collaboration between independent vehicles, facilitated through the use of CQELS queries, will aid in the building of common maps, which can be shared with all vehicles in the flotilla.

\begin{acknowledgement}
This work was funded in part  
funded by the Deutsche Forschungsgemeinschaft (DFG,
German Research Foundation) under the COSMO project (ref. 453130567), the German Ministry for Education and Research as BIFOLD - Berlin Institute for the Foundations of Learning and Data (refs 01IS18025A and 01IS18037A), by the Deutscher Akademischer Austauschdienst (DAAD, ref. 57440921), and by the European Commission H2020 MOSAICrOWN project (ref. 825333).
\end{acknowledgement}
\bibliographystyle{spmpsci}
\bibliography{ref.bib}
\end{document}